%% file: acl_latex.tex
\pdfoutput=1

\documentclass[11pt]{article}

\usepackage[final]{acl}

\usepackage{times}
\usepackage{latexsym}

\usepackage[T1]{fontenc}

\usepackage[utf8]{inputenc}

\usepackage{microtype}

\usepackage{inconsolata}

\usepackage{graphicx}

\usepackage{multirow}
\usepackage{colortbl}
\usepackage{booktabs}
\usepackage{arydshln}
\usepackage[flushleft]{threeparttable}
\usepackage[figuresleft]{rotating}

\title{Do language models accommodate their users? \\ A study of linguistic convergence}

\author{
  \textbf{Terra Blevins\textsuperscript{1,2}} \hspace{0.5cm} \textbf{Susanne Schmalwieser\textsuperscript{3}} \hspace{0.5cm}
  \textbf{Benjamin Roth\textsuperscript{2,3}} 
\\
  \textsuperscript{1}Khoury College of Computer Sciences, Northeastern University \\
  \textsuperscript{2}Faculty of Computer Science, University of Vienna, Vienna, Austria \\
   \textsuperscript{3}Faculty of Philological and Cultural Studies, University of Vienna, Vienna, Austria \\
  \small{
    \textbf{Correspondence:}
    t.blevins@northeastern.edu
  }
}

\begin{document}
\maketitle
\begin{abstract}
While large language models (LLMs) are generally considered proficient in generating language, how similar their language usage is to that of humans remains understudied. In this paper, we test whether models exhibit \textit{linguistic convergence}, a core pragmatic element of human language communication: do models adapt, or converge, to the linguistic patterns of their user? To answer this, we systematically compare model completions of existing dialogues to original human responses across sixteen language models, three dialogue corpora, and various stylometric features. We find that models strongly converge to the conversation's style, often significantly overfitting relative to the human baseline. While convergence patterns are often feature-specific, we observe consistent shifts in convergence across modeling settings, with instruction-tuned and larger models converging less than their pretrained and smaller counterparts. Given the differences in human and model convergence patterns, we hypothesize that the underlying mechanisms driving these behaviors are very different.

\end{abstract}

\section{Introduction}

Large language models have revolutionized natural language generation, with both proprietary and open-source chatbots providing grammatical and topical responses to user queries \cite[e.g.,][]{grattafiori2024llama, team2025gemma}. These models have become so fluent (particularly in English) that readers frequently cannot determine whether a text was authored by a human or a model \cite{clark2021all}. The apparent linguistic competency of LLMs thus opens an array of questions on the properties of model-generated language, particularly concerning how model language use differs from humans'.

In this work, we investigate this through the lens of \textit{accommodation}, the process by which people adjust their speech or writing style based on the identity of their interlocutor \cite{giles1991accommodation}. While accommodation takes many forms, in this paper we focus on \textit{linguistic convergence}, or how similar the target's language is to that of their interlocutor \cite{niederhoffer2002linguistic}. As the most prevalent use case for LLMs is via a chat agent, convergence acts as an interesting case study into the novel setting of human-model interaction. Thus, we seek to answer the following questions: Do LLM-based chatbots adapt their language use to mirror that of their users? And does this behavior mirror trends observed in human accommodation or follow different patterns?

To study convergence in LLMs, we perform a systematic analysis of model-generated responses to human-authored conversations, collected by prompting various LLMs to complete turns in existing conversations drawn from dialogue corpora. We then analyze these responses for a series of stylometric features used to study accommodation in human linguistics, in order to estimate the degree of convergence LLMs exhibit towards their context relative to randomly sampled utterances and a human control (i.e., the gold text in turns that the model replaces).

Our experiments show that models strongly adapt their output style to their context. We also find differences in convergence across model size and training schemes, with larger and instruction-tuned models converging less than smaller or solely pretrained models. Given that pretrained models often \textit{overconverge} on these linguistic markers relative to the human baseline, this suggests that instruction-tuned model responses are often more in line with human behavior. Furthermore, these convergence patterns are feature-specific, with models adapting to their contexts to varying degrees across metrics. 
While this varied behavior across features is also common in human accommodation, the variance we observe in the models often differs from that of the human baseline, indicating subtler differences in human and model convergence. 
We then conclude by discussing the implications of this work towards understanding the linguistics of LLMs.

To summarize, this paper presents the following primary contributions: (1) we develop a novel, synthetic paradigm for testing model behavior with respect to a paired human control, which we apply to studying linguistic convergence in LLMs; and (2) we present a quantitative analysis of how language models converge to the style of their context, with insights into how modeling factors affect this phenomenon and the implications it has for interpreting model-generated text. Our findings demonstrate how models do and do not adapt to their inputs across various stylistic features, providing broader insights into the relationship between human and model language use.\footnote{The source code and model generations can be found at \url{https://github.com/blvns/llm-convergence}.}

\section{Related Work}
Linguistic accommodation has been widely studied in human communication \cite[][inter alia]{giles1991accommodation, niederhoffer2002linguistic, giles2023communication}. Many prior works have leveraged computational methods to characterize this phenomenon in human language across settings ranging from online interactions \cite{mukherjee2012analysis, berdicevskis2023say} to code-switching \cite{bawa2018accommodation} and movie dialogues \cite{danescu2011chameleons}. \citet{bhatt2021detecting} study how users exhibit accommodation when interacting with models, while \citet{parekh2020understanding} examine how users accommodate model code-switching in conversation. In this paper, we apply the method presented in \citet{ireland2011language} to model-generated text while expanding the analysis to include additional stylometric features, to better understand how \textit{models} adapt their language to their users. 

Accommodation research methods have also been applied to computational models of language in prior work. Earlier work has examined the relevance of model accommodation in chatbot task effectiveness \cite{chaves2019s, thomas2020expressions}. Most similar to this paper is \citet{kandra2025llms}, which tests whether GPT-4o exhibits syntactic accommodation. However, \citet{kandra2025llms} focuses on entirely machine-generated interactions between two LLM agents, while our experiments test the extent to which LLMs show accommodation while completing a pre-existing dialogue. Thus, our setting enables direct comparison between human and model responses in a given context. We also evaluate a larger suite of open-source language models across two model families and study a large set of stylometric features to fully characterize the convergence of these models.

Finally, this work also falls into the broader field of language model behavioral analysis (see \citet{chang2024language} for a survey of this area). In particular, our analysis of the stylistic convergence of LMs to their inputs corroborates findings on syntactic and semantic priming in language models \cite{sinclair2022structural, jumelet2024language, gonen2025does}. However, we characterize this convergence more broadly for user-model interactions, rather than through carefully constructed behavioral probes.

\section{Measuring Linguistic Convergence in Language Models}
\label{sec:method}

\textit{Linguistic convergence} occurs when a speaker adapts their language to mirror that of the person they are communicating with. While it is commonly accepted that humans frequently accommodate one another in this manner \cite[e.g.][]{niederhoffer2002linguistic}, it is unknown to what extent language models exhibit these patterns during human-model interaction. We test whether \textit{convergence} occurs in machine-generated text via a two-step process. First, we elicit generations that are grounded in human-authored conversations. We then test whether the model responses exhibit convergence compared to several control settings.

\textbf{Data Generation} First, we elicit machine-generated texts grounded in a dialogue by prompting a model to generate a response $r_t$ continuing a two-person dialogue with speakers $s_x, s_y$, conditioned on the prior turns of the conversation $r_{<t} = r_0,...,r_{t-1}$; the model is always given access to at least $m=5$ prior turns of dialogue. Therefore, the model replaces the utterance of speaker $s_y$ beginning at $t=6$ and continues to participate in the role of $s_y$ on turns $t=\{8, 10,...\}$ for the remainder of the dialogue. 

By replacing utterances within human-authored conversations, our data generation approach approximates a human-model experimental paradigm while minimizing annotator burden. Thus, this setting enables us to extend our experimental framework to many different conversations and domains, beyond what is feasible in a single user study. Our setting also aligns model generations with the original human utterances from conversations, allowing for direct comparisons that are not possible in fully synthetic analyses of LM linguistics.

\textbf{Convergence Analysis} We then quantify how well $r_t$ accommodates other utterances in the dialogue according to a number of different stylometric features (Section \ref{sec:method-stylometrics}). Unless otherwise stated, we consider how much $r_t$ converges with respect to $r_{t-1}$, or the turn immediately prior uttered by the other speaker $s_x$. Thus, we primarily consider specifically how much the model's output mirrors the linguistic features of the most recent utterance from the user; Section \ref{sec:results-stepwise} extends this to a stepwise analysis to examine the effect of earlier turns ($r_{t-2},...$) on model convergence.

\textbf{Baselines} In addition to directly measuring model convergence, we also compare the models' behavior to two baselines: the \textit{human} baseline, which considers how well the original utterance that $r_t$ replaces accommodates the prior text,\footnote{I.e., establishing the accommodation exhibited by humans in the same setting.} and the \textit{random} baseline, which estimates the baseline level of accommodation in a dataset by replacing $r_t$ with $r_{rand}$, an utterance drawn from a random conversion and timestep in the same dataset. \\

\subsection{Linguistic Indicators of Convergence}
\label{sec:method-stylometrics}
We measure the following features to characterize whether models alter their linguistic style to match that of their users, drawing on both human accommodation research and other common stylometric features for linguistic analysis \cite{lagutina2019survey}: 

\textbf{\texttt{Utterance Length}} We measure how similar model response lengths are to the text in prior turns, a feature commonly used in accommodation and stylometric work, such as in \citet{niederhoffer2002linguistic, lin2017stylistic}. We measure this with the symmetric metric from \citet{ireland2011language}: $LSM_x=1-|a-b|/(a+b)$, where $a$ and $b$ represent the observed values for turn $r_t$ and $r_{t-1}$, respectively.

\textbf{\texttt{LIWC} Agreement} A standard measure of linguistic accommodation is the frequency of LIWC (Linguistic Inquiry and Word Count) \cite{chung2012linguistic} function word classes \cite[e.g.][]{danescu2011chameleons}. Here, we consider the LIWC2007 classes considered in \citet{ireland2011language} (personal and impersonal pronouns, articles, conjunctions, prepositions, auxiliary verbs, frequently used adverbs, negations, and quantifiers), also using the $LSM$ metric to calculate how well each LIWC category distribution in generated responses aligns with prior turns. We both report the mean \texttt{LIWC agreement} across categories and provide a fine-grained analysis of each category in Section \ref{sec:results-liwc}.

\textbf{\texttt{PROPN Overlap}} We calculate the overlap (percentage) of proper nouns between the text generated by the model and the preceding turn. We expect that language users who converge more with their interlocutor will have greater overlap in salient semantic topics, as approximated here by proper nouns.

\textbf{\texttt{Token Novelty}} We evaluate the percentage of tokens novel relative to the reference utterance; a lower percentage of novel tokens indicates that the model is converging more towards the user, or exhibiting greater \textit{lexical alignment} \cite{pickering2004toward}. This is measured as $|w_t \cap w_{t-1}|/|w_t|$, where $w_{x} = \{w \in r_x\}$; prior work apply similar metrics of token novelty (or, inversely, token overlap) to measure lexical alignment between interlocutors \cite[i.a.]{ward2007automatically, duplessis2017automatic} . \\


\section{Experimental Setup}
\textbf{Datasets} We perform our dialogue prompting experiments on three popular English datasets: \texttt{DailyDialog} \cite{li2017dailydialog}, containing conversations about daily life as written by English language learners; \texttt{NPR} \cite{majumder2020interview}, a dataset of radio interview transcripts; and the \texttt{Movie} Corpus \cite{danescu2011chameleons}, which contains conversations scraped from movie scripts. Each dataset covers a different dialogue domain; due to their respective data sources, all represent non-spontaneous dialogues (ranging from fully fictional conversations to constrained interviews).

For each dataset, we filter the conversations to ensure they contain at least six turns of dialogue and two speakers; we merge consecutive turns from the same speaker into a single turn. We randomly downsample the larger datasets to consider at most 1,000 conversations per setting. Our experiments are performed on the (filtered) development sets of \texttt{DailyDialog} and \texttt{NPR}; as the \texttt{Movie} corpus does not provide data splits, we randomly sampled our evaluation data from the full set. Table \ref{tab:data-stats} presents the dataset statistics for each corpus post-filtering.

\begin{table}[h]
    \small
    \centering
    \begin{tabular}{l r r r}
    \toprule
        & \multicolumn{3}{c}{\textbf{Dataset Statistics}} \\
        & DailyDialog & Movie & NPR \\
    \hline
    Conversations & 707 & 1,000 & 1,000 \\
    Avg. Turns & 9.79 & 8.98 & 17.57 \\
    Avg. Turn Length & 13.44 & 10.87 & 48.43 \\
    \hdashline
    Replaced Turns & 1,918 & 2,280 & 6,568 \\
    \toprule
    \end{tabular}
    \caption{Dataset sizes and statistics for the dialogue corpora post-filtering. For each dataset, we calculate convergence over the model completions of \textit{Replaced Turns} in each dataset.}
    \label{tab:data-stats}
\end{table}

\textbf{Models} We consider two open-source LLM families: \textit{Gemma 3} \cite{team2025gemma}, with models spanning 1B, 4B, 12B, and 27B parameters; and \textit{Llama 3} \cite{grattafiori2024llama}, with models of 1B, 3B parameters from Llama 3.2 and 8B and 70B from the Llama 3.1 release. We perform inference with checkpoints provided via Huggingface \cite{wolf2019huggingface} and use 8-bit quantization\footnote{\url{https://huggingface.co/docs/bitsandbytes/}} to run the largest model (i.e., Llama3 70B). For each model, we analyze the convergence expressed by both the pretrained and instruction-tuned versions.

\textbf{Prompting}
For each dialogue turn we want the model to complete, we prompt the model to ``Continue this conversation based on the given context'' and provide the conversation history, including prior model generations from earlier turns in the conversation if applicable. We perform simple post-hoc filtering of the generations to remove noise, such as standardizing white space and filtering dialogue tags used within the prompt. Appendix \ref{app:exp-setup} provides example inputs and model generations and other experimental details.

\textbf{Linguistic Annotations} We parse each uttrance with spaCy \cite{honnibal2020spacy} to tokenize the data and obtain proper noun annotations, and we use the LIWC 2007 word classes \cite{chung2012linguistic} to obtain LIWC categories.

\begin{figure}[h]
    \centering
    \includegraphics[width=0.9\linewidth]{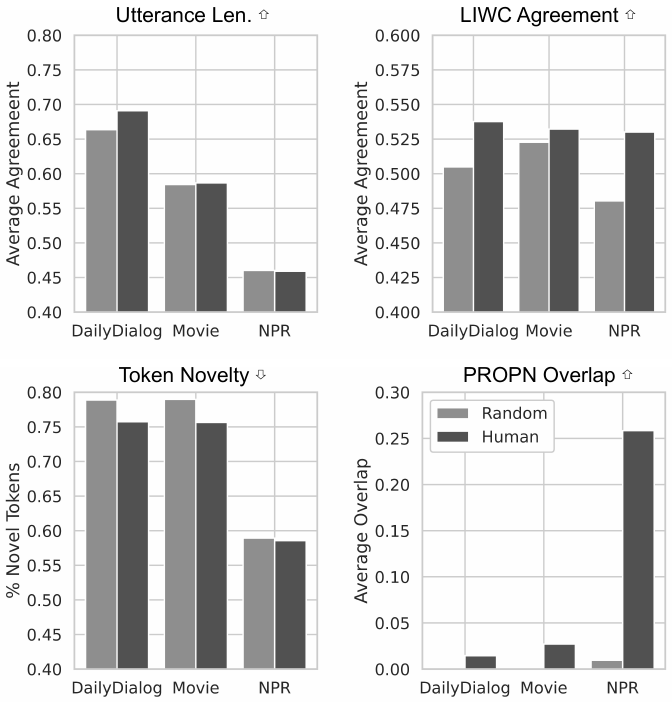}
    \caption{Comparison of the human and random baselines on each metric across datasets. Metrics marked with $\uparrow$ indicate more agreement with higher values; and $\downarrow$, vice-versa.}
    \label{fig:baseline-comp}
\end{figure}

\begin{figure*}[t!]
    \centering
    \includegraphics[width=\linewidth]{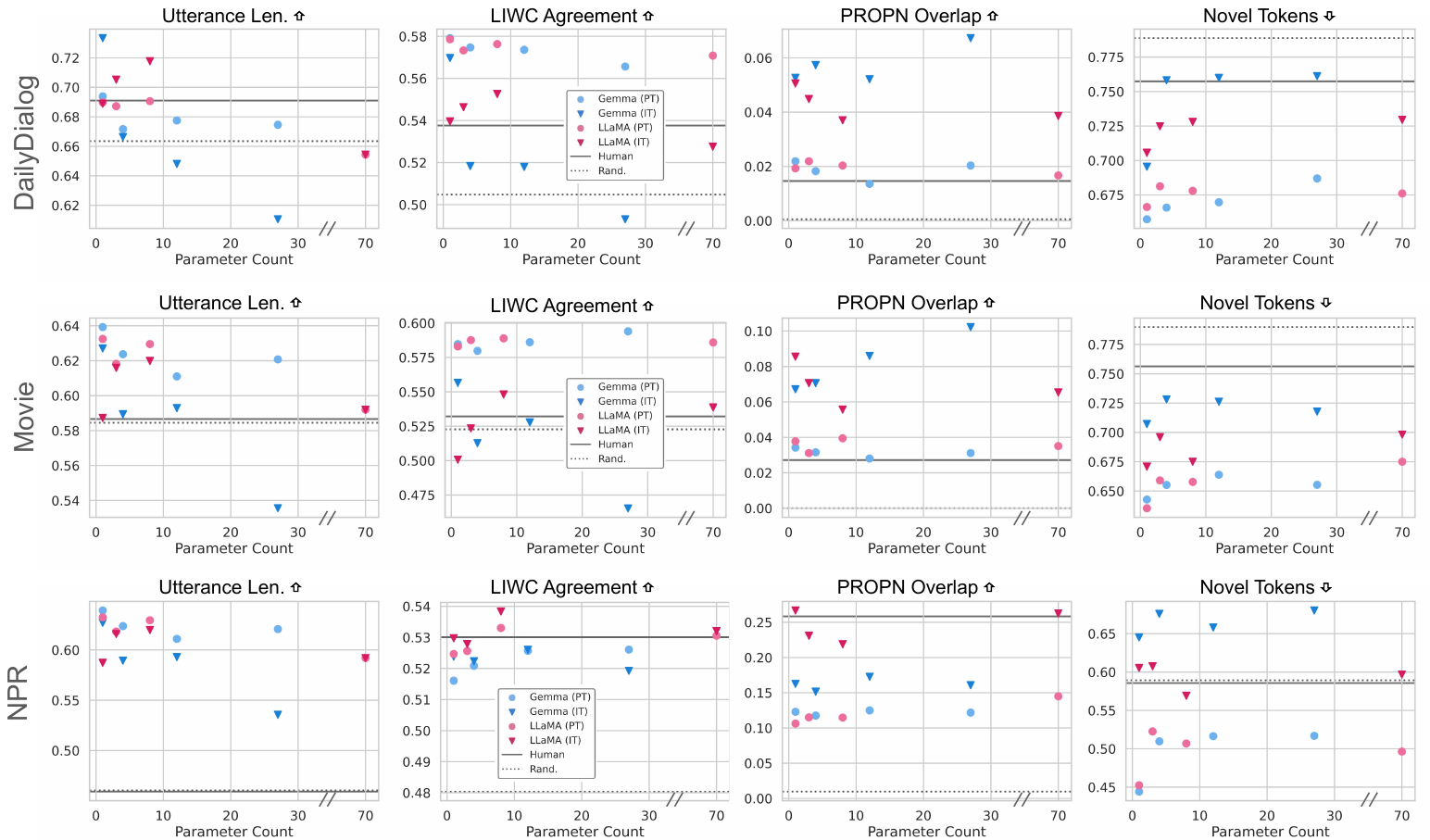}
    \caption{Scatter plot of Gemma and Llama Model scores on various convergence metrics relative to human and random baselines on DailyDialog (top row), Movie corpus (middle), and NPR (bottom), across model sizes (Billion parameters). \textit{PT} indicates pretrained checkpoints while \textit{IT} are instruction-tuned. Metrics marked with $\uparrow$ indicate more agreement with higher values; and $\downarrow$, vice-versa.}
    \label{fig:all-scatter}
\end{figure*}

\section{Analysis Results}
\label{sec:results}

Here, we summarize and discuss the results of the convergence analysis for both the human baselines and model-generated responses when compared to a randomly sampled control.\footnote{Appendix \ref{app:results} presents additional visualizations and the complete set of numerical results.} 
\textbf{LLMs adapt significantly to the style of their interlocutor} across stylometric features, often matching or exceeding the level of convergence exhibited in the human baseline (Figure \ref{fig:all-scatter}).

However, we note that linguistic convergence is often multi-faceted, and human accommodation behavior often varies significantly across features in prior work \cite[e.g.,][]{ireland2011language, danescu2011chameleons}. We find this is also the case for model convergence, particularly in the case of fine-grained linguistic features (Section \ref{sec:results-liwc}) and when examining convergence according to earlier turns in the conversation (Section \ref{sec:results-stepwise}). 
We also observe different trends depending on the model type, with instruction-tuned models generally converging to their context less than their pretrained counterparts.

\subsection{Linguistic Convergence in Human- Authored Text}
\label{sec:results-baseline-analysis}
To obtain a baseline for the expected level of convergence in our chosen dialogue settings, we first examine the linguistic convergence exhibited by the original speakers in these datasets. We therefore compare the level of accommodation (as measured by our four convergence metrics) exhibited in ground truth utterances with \textit{random} utterances drawn from the dataset (Figure \ref{fig:baseline-comp}).

Unsurprisingly, we generally find that the gold utterance $r_t$ converges with the preceding utterance $r_{t-1}$ more than a randomly sampled utterance. We find a significant difference ($p < 0.05$ in a paired t-test) between the \textit{human} and \textit{random} settings on the \texttt{token novelty} and \texttt{PROPN overlap} metrics across datasets. However, convergence in \texttt{utterance length} is only significant between the two settings on the DailyDialog dataset. 

We also observe differing levels of convergence on \texttt{LIWC} categories: while NPR conversations exhibit significant accommodation over the random baseline on all LIWC categories except \textit{quantifiers}, the other datasets show much smaller differences.\footnote{See Appendix Tables \ref{tab:liwc-results-dailydialog}, \ref{tab:liwc-results-movie}, \ref{tab:liwc-results-npr} for full numerical results.} DailyDialog conversations only show significant differences on five of nine \texttt{LIWC} word classes (despite appearing to have a large average increase in convergence over random utterances), and utterances drawn from the Movie corpus are only significantly more accommodating over the random baseline on a single class, \textit{auxiliary verbs}.

The limited linguistic convergence we observe in the Movie corpus is likely due to the nature of the data, as the conversations were written by screenwriters, rather than drawn from naturally occurring speech. \citet{danescu2011chameleons} similarly found that while the Movie corpus exhibited convergence under their conditions, the levels observed were much lower compared to real-world conversations (in their case, drawn from Twitter).

\subsection{Linguistic Convergence in LLM- Generated Text}
\label{sec:results-model-analysis}
We now turn to examining how model-generated responses to these conversations exhibit convergence. Figure \ref{fig:all-scatter} compares model scores on each convergence metric against the human and random baselines on the three datasets. With these results, we consider the following questions:

\paragraph{Do models converge to their context?}
We find that models significantly outscore the random baseline in 81.25\%, 100\%, and 85.42\% of cases for \texttt{Utterance Length}, \texttt{PROPN overlap}, and \texttt{Token Novelty}, respectively, in a paired t-test (p < 0.05). We also observe strong convergence on LIWC categories: models outscore the random utterances 91.67\% (44 of 48) of the time on averaged scores, with significant improvements on individual LIWC classes ranging from 37.5\% (for conjunction words) to 87.5\% (on personal pronouns). While specific model convergence trends can differ based on several factors, this indicates that models generally adapt to the linguistic style of their context. 

Furthermore, language models also significantly \textit{over-converge} relative to human-authored utterances in many cases. We find that in 62.5\% and 79.2\% of cases for \texttt{Utterance Length} and \texttt{Token Novelty}, the model-generated responses significantly outscore human utterances. Overconvergence occurs in fewer cases for \texttt{PROPN overlap} (35.4\%) and LIWC classes, where they range between 10.4\% and 56.25\% of settings. We discuss \textit{which} model settings overfit to their context relative to humans below.

\paragraph{Does model size and training affect convergence?}
Figure \ref{fig:all-scatter} shows that model training approaches affect the convergence behaviors of LLMs, with pretrained models generally adapting more to their context than their instruction-tuned counterparts. This is particularly true in the case of the Gemma model family, where the instruction-tuned models exhibit the least amount of convergence. An exception to this trend is the \texttt{PROPN Overlap} metric: here, pretrained models more closely mirror the human baseline while
while instruction-tuned models significantly \textit{over-accommodate} by more often repeating proper nouns from the prior utterance.
This difference is likely due to alternate training objectives; for example, pretrained models likely \textit{appear} to adapt more on \texttt{Token Novelty} because they are trained to fit closely to the input distribution, while instruction-tuned models are encouraged to introduce novel information during fine-tuning.

We also see minor convergence trends across model size: larger models slightly but nonsignificantly shift towards the human baseline and accommodate their context less on \texttt{Utterance Length}, \texttt{LIWC Agreement}, and \texttt{PROPN Overlap} (Appendix Table \ref{tab:app-model-size}). However, convergence trends appear more stable for the \texttt{Token Novelty} metric.

\begin{figure}[b!]
    \centering
    \includegraphics[width=\linewidth]{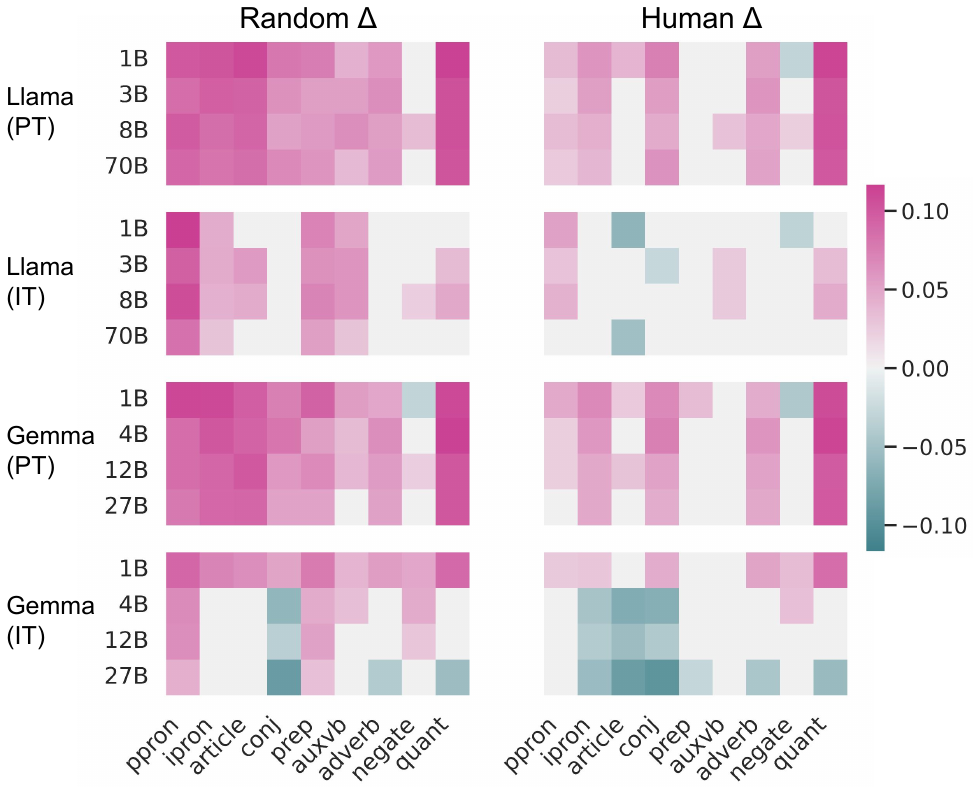}
    \caption{Summary of model convergence relative to the human and random baselines for individual LIWC word classes on the \textit{DailyDialog} dataset. \textcolor[RGB]{200,74,148}{Pink} cells indicate classes where the model significantly ($p<0.05$) overconverges relative to the baseline, while \textcolor[RGB]{72, 131, 141}{green} cells indicate significant undercongergence. Gray cells are not significantly different from the baseline.
    \vspace{-5pt}}
    \label{fig:dailydialog-liwc}
\end{figure}

\paragraph{Does model convergence differ across datasets?}
We find that LLMs exhibit relatively consistent convergence behavior across the three datasets, frequently significantly outscoring the random control. Given the presence of convergence across datasets and indicator metrics, linguistic convergence appears to be a general phenomenon in LLMs across various data settings and model types.

However, we also observe shifts in the models' adaptation to different datasets, particularly in the case of the \textit{NPR} dataset. Specifically, models often exhibit \textit{less} convergence than the baselines on NPR for \texttt{Token Novelty} and \texttt{PROPN Overlap}, but significantly more convergence in terms of \texttt{Utterance Length} (Figure \ref{fig:all-scatter}). This shift likely stems from the underlying data: compared to the more casual dialogue of our other settings, \texttt{NPR} conversations are interview transcripts that have more structured turns and expected variance in utterance lengths, particularly between the interviewer and interviewee. Thus, while models adapt their style to their contexts, whether this adaptation is \textit{human-like} depends on the setting. 

\begin{figure*}
    \centering
    \includegraphics[width=1\linewidth]{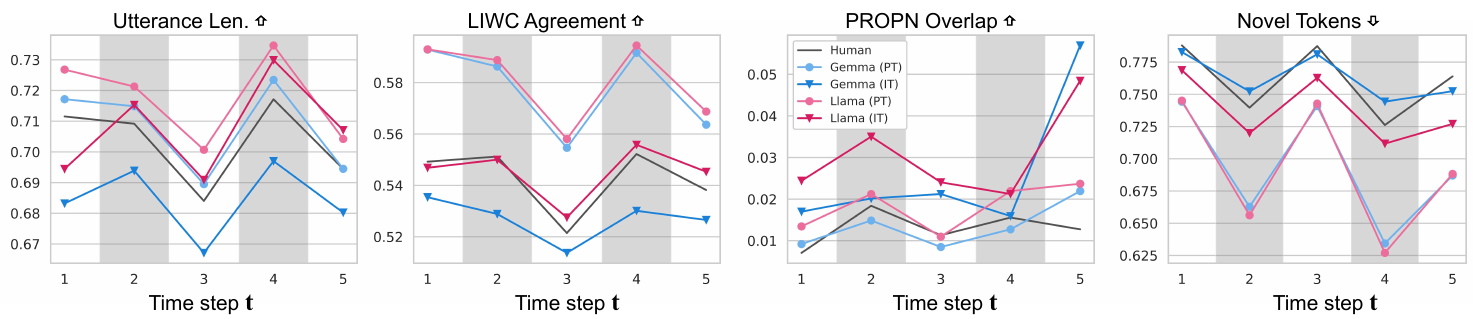}
    \caption{Stepwise analysis of convergence in LM generations (and human ground truth utterances) for \texttt{DailyDialog}, measuring the agreement between each utterance $r_{t=n}$ and the preceding utterances $r_{t=1,...,n-1}$ on our four metrics. Timesteps in gray ($t=2,4$) indicate the prior turns in the role the model adopts, $S_y$, while white timesteps are utterances from the other speaker, $S_x$. Each line reports the averaged score across all model sizes in a given family.}
    \label{fig:dailydialog-stepwise}
\end{figure*}

\subsection{Fine-grained Analysis of LIWC Categories}
\label{sec:results-liwc}

While we examine several axes to quantify linguistic convergence in language models, the majority of prior computational work on human accommodation quantifies convergence by measuring the frequency of common types of function words based on the subset of \texttt{LIWC} word classes identified by \citet{ireland2011language}.
In this section, we consider model convergence on these nine \texttt{LIWC} sub-categories (rather than the averaged result across classes reported in prior sections) to understand finer-grained aspects of when models do and don't converge to their context.

Figure \ref{fig:dailydialog-liwc} summarizes the results of our fine-grained LIWC analysis on DailyDialog, reporting the relative delta of model scores on these classes compared to the random and human baselines; Appendix Figures \ref{fig:movie-liwc} and \ref{fig:npr-liwc} present the results for the \textit{Movie} and \textit{NPR} datasets. We see similar general trends to previous sections (e.g., instruction-tuned models converge less to their context than pretrained ones), but model behavior often varies markedly across individual \texttt{LIWC} categories, particularly in comparison to human scores. Specifically, while pretrained model convergence is usually stronger than the random baseline, their convergence relative to the human baseline is mixed. Compared to human utterances, the pre-trained models overconverge on certain word classes, such as quantifiers, conjunctions, and adverbs, while performing on par with humans on others. 

A potential explanation is that pretrained models may simply generate these word classes less overall. Examining the use of quantifiers by the largest pretrained Llama3 model, we find it underuses this word class relative to the human texts at similar rates, regardless of whether the prior utterance includes at least one quantifier or not (28.6\% vs. 24.1\%). Table \ref{tab:liwc-qual} (Case A) shows a pair of model- and human-generated utterances exemplifying this trend for quantifiers.
Given that only 27\% of the reference utterances include a quantifier, it is likely the ``overconvergence'' we see is at least partially due to overall lower use of uncommon word types by the pretrained models.

Furthermore, instruction-tuned models continue to converge less than the pretrained ones on individual \texttt{LIWC} classes, underconverging relative to human utterances in 16 cases. Here we see a strong difference across model families, with the instruction-tuned Llama models performing similarly to humans on most word classes and their Gemma counterparts underconverging on a number of categories, such as impersonal pronouns, articles, and conjunction usage. There is also a minor shift with model size, as larger Gemma models converge less across these classes. 

In contrast with pretrained models, this underconvergence seems to stem from an \textit{overuse} of these classes, which causes generations from the instruction-tuned models to align less with their context. We examine the use of conjunctions by the pretrained and instruction-tuned largest Gemma models, and find that instruction-tuned models frequently overgenerate conjunctions relative to the human and pretrained model, with instruction-tuned models using additional conjunctions in 47.0\% and 59.1\% of utterances, respectively (for reference, the human reference only uses this word class in 39\% of utterances). Table \ref{tab:liwc-qual} (Case B) provides an example of this overuse.

\begin{table}[]
    \centering
    \small
    \begin{tabular}{l l}
        \toprule
        Case A (Quant.) \\
        Prior Utterance & ``I will take ten.''\\
        \hdashline
        Human & ``Do you want to add \textbf{some} baby's \\
         & breath for that?''\\
        Llama-70B (PT) & ``Do you want them delivered?'' \\
        \midrule
        Case B (Conj.) \\
        Prior Utterance & ``Yes, our over-the-counter medicine \\
        & is over there on that shelf.''\\
        \hdashline
        Gemma-27B (PT) & ``Thank you.'' \\
        Gemma-27B (IT) & ``Great. \textbf{And} how much is a bottle \\
        & of 200?'' \\
        \bottomrule
    \end{tabular}
    \caption{Qualitative examples of model behavior on fine-grained stylometric features, drawn from DailyDialog experiments. \textbf{Case A}: Pretrained models overconverge on \textit{quantifiers}; \textbf{Case B}: instruction-tuned models overuse (and thus underconverge on) \textit{conjunctions}.
    \vspace{-9pt}} 
    \label{tab:liwc-qual}
\end{table}

\subsection{Stepwise Analysis of Linguistic Convergence}
\label{sec:results-stepwise}
In the previous analyses, we examine how well the model-generated responses $r_t$ converge linguistically to the immediately prior utterance in the conversation $r_{t-1}$. However, our experimental setup primes the model with the first five turns in a dialogue before prompting it to participate. Thus, there is much more context for the model to potentially fit to than $r_{t-1}$, as it is also conditioned on other prior turns from $S_x$, the speaker with whom the model is asked to converse (and the standard target of linguistic convergence analysis), and on turns from $S_y$, the speaker that the model replaces.

In this section, we test how linguistic features from earlier turns in the conversation influence initial model-generated utterances in conversation. Specifically, we take the first utterance generated by the model in each conversation (at $t=6$) and compare it to all prior turns $r_{t=1,...,5}$ (Figure \ref{fig:dailydialog-stepwise} for DailyDialog; Appendix Figures \ref{fig:movie-stepwise} and \ref{fig:npr-stepwise} for Movie and NPR datasets). For clarity, we report average convergence scores across model sizes on each model family and training scheme set.

We find that human convergence scores fluctuate across timesteps; this is unsurprising, as alternating timesteps are uttered by the same speaker $S_y$ as the considered utterance $r_t$, indicating that each speaker's linguistic patterns agree more with themselves than with their interlocutors. Interestingly, we see similar patterns across time with the language models for most metrics. This finding suggests that not only do language models adapt to their context, they also differentiate this adaptation to the particular speaker role \cite{sacks1974simplest} they are replacing within the dialogue.\footnote{Interlocutors generally take different roles (speaker, listener) to coordinate turn-taking in conversation.} While these models demonstrate sensitivity to speaker roles when emulating existing speakers, it remains an open question whether they maintain distinct speaker role patterns in open-ended dialogues without a specific conversational partner to emulate.

However, we do observe some differences in model behavior compared to humans across time, particularly on \texttt{PROPN Overlap}. Model behavior follows the human trend less closely for this metric at earlier timesteps, with a sharp increase in overlap scores relative to humans on the last turn before the model generation (Figure \ref{fig:dailydialog-stepwise}). This increased score corresponds to the overconvergence observed in models in the prior sections on \texttt{PROPN}. Thus, it is likely that in the case of exact word overlap (e.g., names and other proper nouns), model convergence demonstrates a strong \textit{recency bias} towards newer concepts \cite{liu2024lost}.\footnote{We similarly see small upticks on convergence with the first utterance for multiple datasets and metrics, suggesting that \textit{primacy biases} may affect convergence as well.}

\section{Discussion}
\label{sec:discussion}

Throughout our analysis, we compare model convergence to that of the human baselines to contextualize the models' behavior. 
However, while in some cases the model converges to its context similarly to human utterances, we emphasize that the observed similarities do not necessarily indicate that the underlying causes of these behaviors are the same. Human accommodation is driven by speakers (often unconsciously) altering their speech to foster social and communicative goals \cite{giles1991accommodation}. In contrast, language models do not have these same underlying psychological communication goals when generating text.

We hypothesize that model convergence is instead driven by their pretraining objective, which encourages the model to produce test stylistically consistent with their input by training them on complete, often single-author documents. This consistency effect extends beyond style convergence: model generations have also been shown to be influenced by their input in structural priming \cite{sinclair2022structural, jumelet2024language} and through superfluous semantic correlations \cite{gonen2025does}. Recently, \citet{kandra2025llms} demonstrated this in the syntactic convergence of model-model interactions. The stylistic convergence and overconvergence that we observe are thus likely another facet of this behavior.

Instead, an important consideration is how the user will interpret texts from models that \textit{appear} to accommodate them as a human interlocutor would, such as in the case of the ``human-like" behavior we observe in instruction-tuned models. Appropriate model style has been shown to facilitate successful chatbot interactions \cite{chaves2019s, thomas2020expressions}, and \citet{bhatt2021detecting} finds that users tend to accommodate models more when they successfully produce topically relevant outputs, treating them more like human conversation partners. Other work has observed that while sycophantic behavior in langauge models can lead to lower model trust if detected by the user \cite{carro2024flattering}, users who don't identify sycophancy in model generations will often overrely on these unhelpful model responses \cite{bo2025invisible}. Thus, model convergence that better aligns with human behavior will likely mislead users to trust these models more, even if a model is unreliable for a given task in practice.

\section{Conclusion}
This paper presents a comprehensive description of \textit{linguistic convergence} in a series of open-source generative language models. Specifically, we characterize the extent to which these language models adapt their outputs to the style of their context across various stylometric features and dialogue domains. While our experiments reveal varied trends across datasets and model training regimes, we generally find that LM generations \textit{do} exhibit convergence, almost always significantly outscoring a random control on the considered metrics. Furthermore, in many cases, language models also overconverge relative to the human baseline.

Thus, we consider the extent to which model convergence is even related to human accommodation. Our finer-grained analysis of \texttt{LIWC} categories (a prototypical feature for studying accommodation) shows that models exhibit very different patterns on these features than humans, suggesting that the underlying mechanisms for these behaviors are likely very different. Therefore, while this work primarily considers human behavior as a reference for characterizing model generations, future work should further characterize these differences and examine the underlying causes of the observed convergence in LMs.

\section*{Limitations}
We approximate LLM participation in user-driven dialogues by having them complete turns in existing dialogue datasets. While this approach has experimental advantages (e.g., we can directly compare human accommodation features with the model's responses in the same context), it also presents some limitations. Specifically, the model's responses may be biased because it participates only in the later portion of (fixed) conversations; these conversational structures are also likely to differ significantly from those in natural interactions between a user and an LLM. Future work should supplement our experimental setting with more focused user studies to validate whether our findings hold in these cases. We also note that, though we perfrom our experiments on three popular, high-quality English dialogue datasets, each dataset's source data imposes constraints on natural conversation structures, potentially limiting generalization of our findings.

While we perform experiments on sixteen language models, it remains an open question how larger models ($>$70B parameters) and models post-trained on other objectives adapt to their users. The considered datasets cover various styles of conversations, but these differ from how some users interact with the models (i.e., information-seeking). Thus, future work should confirm whether these findings hold up in interactive user studies. Finally, while we test for an array of stylometric features, the model's behavior of other aspects of style may differ.

\section*{Acknowledgments} This research was funded by the
Vienna Science and Technology Fund (WWTF)[10.47379/VRG19008] ”Knowledge infused Deep Learning for Natural Language Processing”. We thank Agnieszka Gwizdek and Antonio Innocenti for early investigation in this space and Lukas Thoma for feedback on this work.

\bibliography{custom}

\appendix

\section{Further Experimental Details}
\label{app:exp-setup}

In this section, we provide additional details about our experimental setup. To obtain model completions for the target conversations, we generate responses by prompting the model to continue the conversation based on the given context (Table \ref{tab:input-example}). We generate from the models using a temperature of 0.4 and top-p sampling at 0.8. We also limit the generated output to 40 tokens and perform simple cleaning heuristics to standardize the generation before performing our convergence analysis. These parameters were chosen after a small parameter search over a small subset of the Movie corpus conversations.

\paragraph{Computational Hardware and Budget}
We run our experiments on 2 H100 GPUs, using 8-bit quantization for the LLaMA3 70B parameter model; in total, across all models (16) and datasets (3), we perform 48 generation runs of up to 1000 conversations (see Table \ref{tab:data-stats} for full dataset composition details). Post-hoc analysis of the data is not computationally expensive and performed on CPUs, using the spaCy ``en\_core\_web\_trf'' model \cite{honnibal2020spacy} for tokenization and proper noun identification.

\paragraph{Artifact Licensing and Use}
The DailyDialog dataset \cite{li2017dailydialog} is released under a CreativeCommons Attribution and Non-comerical license (CC BY-NC-SA 4.0), while the Cornell Movie corpus \cite{danescu2011chameleons} and NPR Interview dataset \cite{majumder2020interview} are released in conjunction with academic papers but do not report the license for these datasets in their papers or associated artifacts. All datasetd are released through academic publications with the intention of contiued use in NLP and dialogue research. 

The spaCy package \cite{honnibal2020spacy} is released under the MIT License (allowing both academic and commercial use), while the LIWC package \cite{chung2012linguistic} is under a custom end user liscense agreement for academic use.\footnote{https://www.liwc.app/help/eula} The Llama3 models \cite{grattafiori2024llama} are released by Meta for open use under a custom liscense\footnote{https://www.llama.com/llama3/license/}, and the Gemma model family is released under the Gemma Terms of Use.\footnote{https://ai.google.dev/gemma/terms}

\begin{table}[]
    \centering
    \footnotesize
    \begin{tabular}{l l}
        \toprule
        Prompt & Continue this conversation based on the \\ 
        & given context. \\
        Context & <user> What can I do for you, sir? </user> \\
         & <assistant> We'd like to order breakfast for \\
         & tomorrow morning. </assistant> ... \\
         & <user> OK, and when shall I bring it here? \\
         & </user> \textbackslash n <model> \\
        \hline
        Responses & \\
        \hdashline
        Human & About seven thirty. By the way... \\[.05cm]
        Generation$^*$ & At 7:30 AM. \\
        \toprule
    \end{tabular}
    \caption{Prompt, context, and sample generations for our prompting setup to obtain model responses for our convergence analysis. $^*$Example text generated by the Llama3 (3B) pretrained model, example conversation drawn from the DailyDialog development set.}
    \label{tab:input-example}
\end{table}

\begin{figure}
    \centering
    \includegraphics[width=\linewidth]{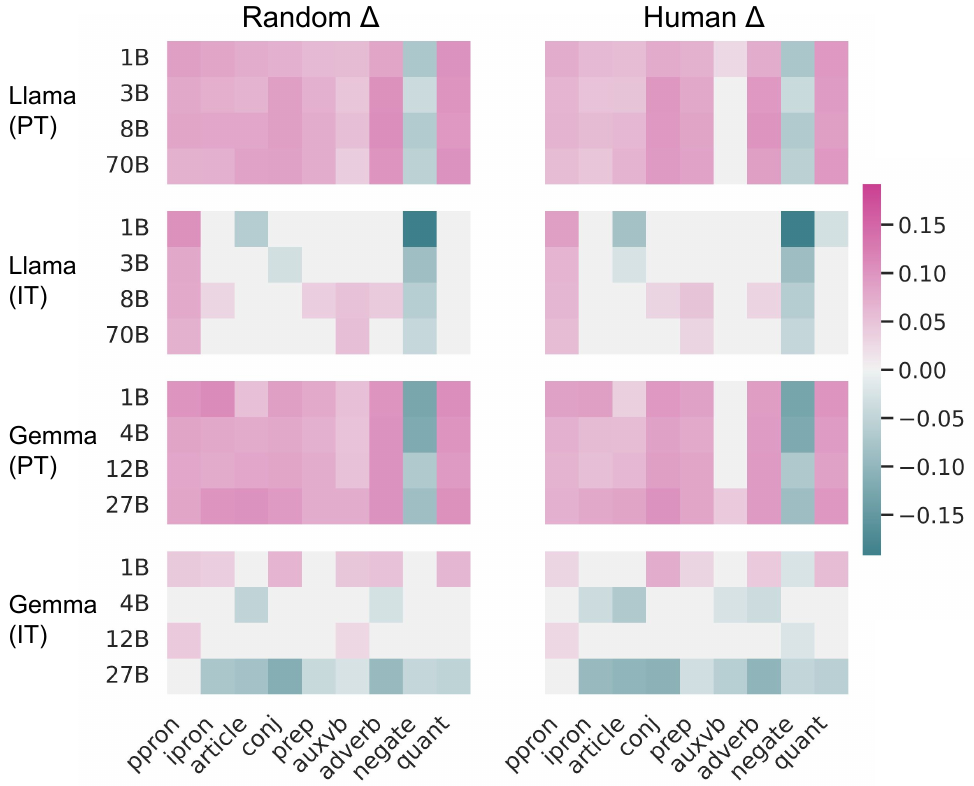}
    \caption{Summary of model convergence relative to human and random baselines on LIWC word classes for \textit{DailyDialog}.}
    \label{fig:movie-liwc}
\end{figure}

\begin{figure}
    \centering
    \includegraphics[width=\linewidth]{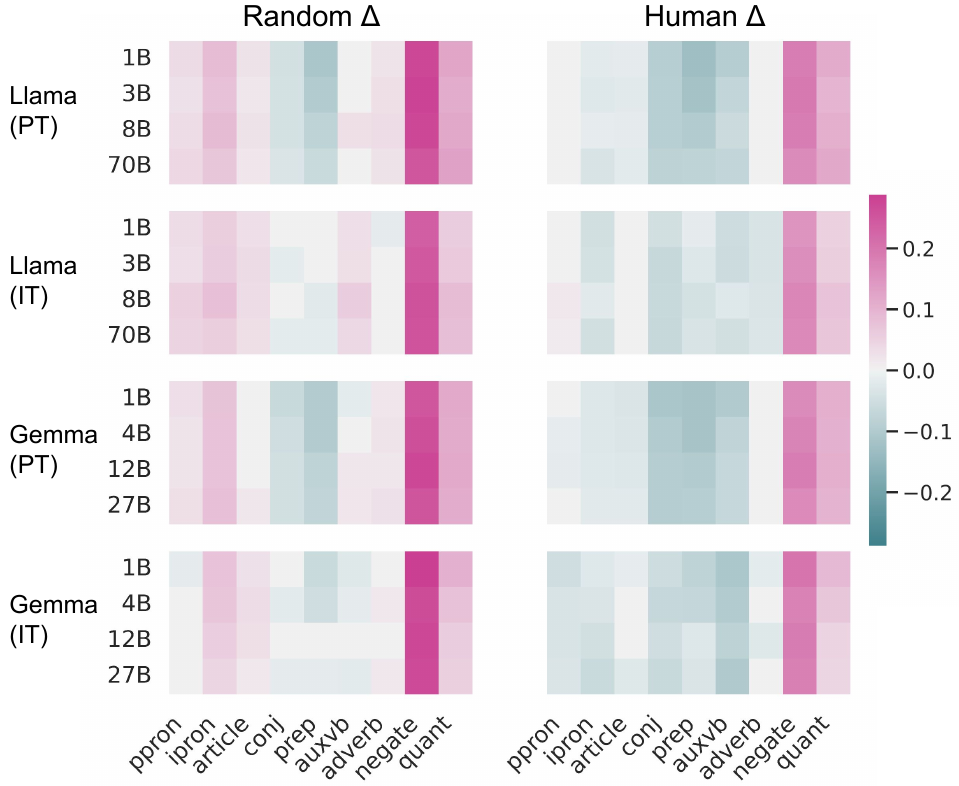}
    \caption{Summary of model convergence relative to human and random baselines on LIWC word classes for \textit{DailyDialog}.}
    \label{fig:npr-liwc}
\end{figure}

\begin{figure*}
    \centering
    \includegraphics[width=1\linewidth]{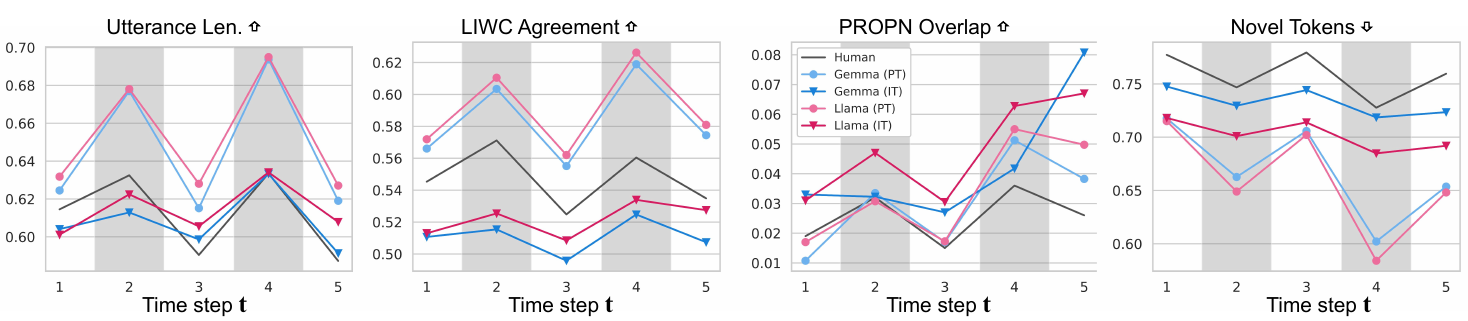}
    \caption{Stepwise analysis of convergence in LM generations (and human ground truth utterances) for the \texttt{Movie} corpus, measuring the agreement between each utterance $r_{t=n}$ and the preceding utterances $r_{t=1,...n-1}$ on our four metrics. Timesteps in gray ($t=2,4$) indicate the prior turns in the role the model adopts, $S_y$, while white timesteps are utterances from the other speaker $S_x$.}
    \label{fig:movie-stepwise}
\end{figure*}

\begin{figure*}
    \centering
    \includegraphics[width=1\linewidth]{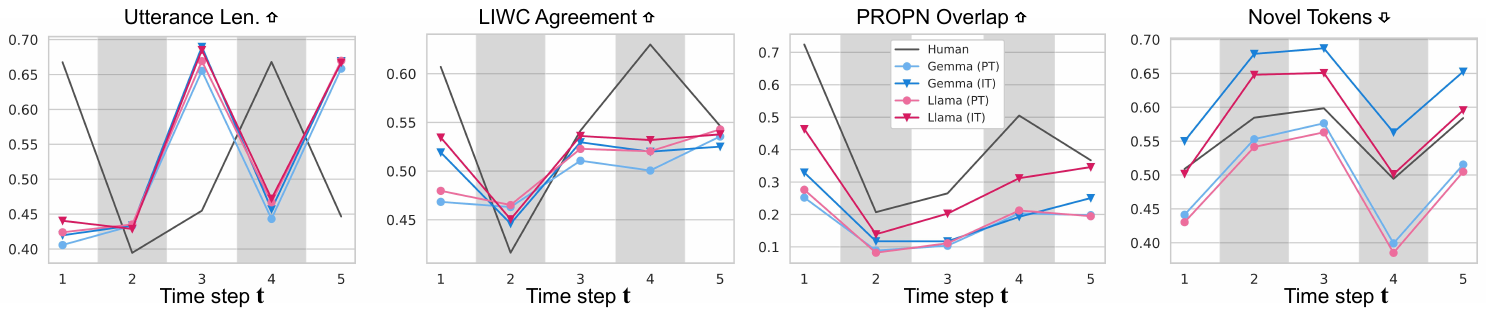}
    \caption{Stepwise analysis of convergence in LM generations (and human ground truth utterances) for \texttt{NPR}, measuring the agreement between each utterance $r_{t=n}$ and the preceding utterances $r_{t=1,...n-1}$ on our four metrics. Timesteps in gray ($t=2,4$) indicate the prior turns in the role the model adopts, $S_y$, while white timesteps are utterances from the other speaker $S_x$.}
    \label{fig:npr-stepwise}
\end{figure*}

\section{Additional Analysis Results}
\label{app:results}

Here, we present additional analysis figures and the full numerical results of our convergence experiments. First, we perform a correlation study to examine the effect of model size and style convergence (Table \ref{tab:app-model-size}).

Figure \ref{fig:movie-liwc} and Figure \ref{fig:npr-liwc} show the relative delta against the human and random baselines for individual \texttt{LIWC} categories for the Movie corpus and NPR dataset, respectively; this complements the DailyDialog results presented in Section \ref{sec:results-liwc}. We also present the full numerical results for the individual \texttt{LIWC} classes for DailyDialog (Table \ref{tab:liwc-results-dailydialog}), the Movie corpus (Table \ref{tab:liwc-results-movie}), and NPR (Table \ref{tab:liwc-results-npr}). 
We also report the summary figures for the stepwise experiments described in Section \ref{sec:results-stepwise} with Figure \ref{fig:movie-stepwise} for the Movie corpus and Figure \ref{fig:npr-stepwise} for NPR. 

Finally, we present the full dataset-level experimental results across our convergence metrics in Tables \ref{tab:app-results-dailydialog}, \ref{tab:app-results-movie}, and \ref{tab:app-results-npr} for the \texttt{DailyDialog}, \texttt{Movie}, and \texttt{NPR} datasets, respectively. We also provide \textit{Human} and \textit{Random} baselines in each table for comparison.

\begin{table}[]
    \small
    \centering
    \begin{tabular}{c c r r}
        \toprule
         Dataset & PT/IT & $\rho$ & $p$  \\
         \hline
         Utterance Length & \\
         \hdashline
         DailyDialog & PT & -0.189 & 0.653 \\
         DailyDialog & IT & -0.221 & 0.380 \\
         Movie & PT & -0.252 & 0.546 \\
         Movie & IT & -0.253 & 0.546 \\
         NPR & PT & -0.263 & 0.530 \\
         NPR & IT & -0.263 & 0.529 \\
         \hline
         LIWC Agreement & \\
         \hdashline
         DailyDialog & PT & -0.225 & 0.592 \\
         DailyDialog & IT & -0.233 & 0.579 \\
         Movie & PT & -0.212 & 0.614 \\
         Movie & IT & -0.256 & 0.541 \\
         NPR & PT & -0.217 & 0.606 \\
         NPR & IT & -0.218 & 0.604 \\
         \hline
         PROPN Overlap & \\
         \hdashline
         DailyDialog & PT & -0.355 & 0.389 \\
         DailyDialog & IT & -0.246 & 0.558 \\
         Movie & PT & -0.246 & 0.557 \\
         Movie & IT & -0.136 & 0.749 \\
         NPR & PT & -0.563 & 0.146 \\
         NPR & IT & -0.282 & 0.499 \\
         \hline
         Token Novelty & \\
         \hdashline
         DailyDialog & PT & -0.209 & 0.620 \\
         DailyDialog & IT & -0.214 & 0.612 \\
         Movie & PT & -0.239 & 0.569 \\
         Movie & IT & -0.230 & 0.584 \\
         NPR & PT & -0.199 & 0.636 \\
         NPR & IT & -0.217 & 0.606 \\
         \toprule
    \end{tabular}
    \caption{Correlation across model size and convergence values for different features and experimental settings.}
    \label{tab:app-model-size}
\end{table}

\input{dailydialog_table}
\input{movies_table}
\input{npr_table}

\input{liwc_dailydialog_table}
\input{liwc_movie_table}
\input{liwc_npr_table}

\end{document}

%% file: dailydialog_table.tex
\begin{table*}[]
    \centering
    \begin{threeparttable}
    \begin{tabular}{l r r r r r r}
    \toprule
       \multirow{2}{*}{{Model}} & \multicolumn{4}{c}{{Accommodation Features}} \\
       & Utterance Len. ($\uparrow$) & LIWC ($\uparrow$) & PROPN ($\uparrow$) & Token Novelty ($\downarrow$) \\
       \hline
       Human & 0.69102 & 0.53759 & 0.01460 & 0.75742 \\
       Random & 0.66348 & 0.50481 & 0.00052 & 0.78874 \\
       \arrayrulecolor{black}\hdashline
       \rowcolor{gray!20} Llama3-1B (PT) & 0.68975 & 0.57859 & 0.01929 & \textbf{0.66624}\tnote{$\dagger$} \\
       \rowcolor{gray!20} Llama3-3B (PT) & 0.68722 & 0.57327 & 0.02190 & \textbf{0.68130}\tnote{$\dagger$}  \\
       \rowcolor{gray!20} Llama3-8B (PT) & 0.69062 & 0.57630 & 0.02033 & \textbf{0.67796}\tnote{$\dagger$} \\
       \rowcolor{gray!20} Llama3-70B (PT) & \textbf{0.65438}\tnote{$\dagger$} & 0.57083 & 0.01668 & \textbf{0.67603}\tnote{$\dagger$} \\
       Llama3-1B (IT) & 0.68893 & 0.53960 & \textbf{0.05057}\tnote{$\dagger$} & \textbf{0.70558}\tnote{$\dagger$} \\
       Llama3-3B (IT) & \textbf{0.70523} & 0.54633 & \textbf{0.04484}\tnote{$\dagger$} & \textbf{0.72480}\tnote{$\dagger$} \\
       Llama3-8B (IT) & \textbf{0.71776}\tnote{$\dagger$} & 0.55260 & \textbf{0.03702}\tnote{$\dagger$} & \textbf{0.72800}\tnote{$\dagger$} \\
       Llama3-70B (IT) & \textbf{0.65438}\tnote{$\dagger$} & 0.52757 & \textbf{0.03858}\tnote{$\dagger$} & \textbf{0.72950}\tnote{$\dagger$}  \\
       \hdashline
       \rowcolor{gray!20} Gemma3-1B (PT) & 0.69385 & 0.57911 & 0.02190 & \textbf{0.65732}\tnote{$\dagger$} \\
       \rowcolor{gray!20} Gemma3-4B (PT) & \textbf{0.67167} & 0.57470 & 0.01825 & \textbf{0.66582}\tnote{$\dagger$} \\
       \rowcolor{gray!20} Gemma3-12B (PT) & \textbf{0.67753} & 0.57359 & 0.01356 & \textbf{0.66968}\tnote{$\dagger$} \\
       \rowcolor{gray!20} Gemma3-27B (PT) & \textbf{0.67461} & 0.56564 & 0.02033 & \textbf{0.68695}\tnote{$\dagger$} \\
       Gemma3-1B (IT) & \textbf{0.73334}\tnote{$\dagger$} & 0.56969 & \textbf{0.05266}\tnote{$\dagger$} & \textbf{0.69551}\tnote{$\dagger$} \\
       Gemma3-4B (IT) & \textbf{0.66628}\tnote{$\dagger$} & 0.51836 & \textbf{0.05735}\tnote{$\dagger$} & 0.75824 \\
       Gemma3-12B (IT) & \textbf{0.64808}\tnote{$\dagger$} & 0.51796 & \textbf{0.05214}\tnote{$\dagger$} & 0.75992 \\
       Gemma3-27B (IT) & \textbf{0.61049}\tnote{$\dagger$} & 0.49306 & \textbf{0.06726}\tnote{$\dagger$} & 0.76123 \\
     \toprule
    \end{tabular}
    \caption{Metric scores of common indicators of linguistic convergence in model-generated responses to conversations in the \textit{DailyDialog} dataset. Scores for \texttt{Utterance Length}, \texttt{PPROPN}, and \texttt{Token Novelty} in \textbf{bold} are significantly different from human metrics ($p<0.05$ on a paired t-test); scores indicated with $\dagger$ are $p < 0.001$ relative to human scores.}
    \label{tab:app-results-dailydialog}
    \end{threeparttable}
\end{table*}

%% file: movies_table.tex
\begin{table*}[]
    \centering
    \begin{threeparttable}
    \begin{tabular}{l r r r r r r}
    \toprule
       \multirow{2}{*}{{Model}} & \multicolumn{4}{c}{{Accommodation Features}} \\
       & Utterance Len. ($\uparrow$) & LIWC ($\uparrow$) & PROPN ($\uparrow$) & Token Novelty ($\downarrow$) \\
       \hline
       Human & 0.58664 & 0.53215 & 0.02719 & 0.75639 \\
       Random & 0.58447 & 0.52271 & 0.00000 & 0.78985  \\
       \arrayrulecolor{black}\hdashline
       \rowcolor{gray!20} Llama3-1B (PT) & \textbf{0.63248}\tnote{$\dagger$} & 0.58302 & 0.03772 & \textbf{0.63531}\tnote{$\dagger$} \\
       \rowcolor{gray!20} Llama3-3B (PT) & \textbf{0.61824}\tnote{$\dagger$} & 0.58764 & 0.03114 & \textbf{0.65910}\tnote{$\dagger$} \\
       \rowcolor{gray!20} Llama3-8B (PT) & \textbf{0.62951}\tnote{$\dagger$} & 0.58892 & \textbf{0.03947} & \textbf{0.65774}\tnote{$\dagger$} \\
       \rowcolor{gray!20} Llama3-70B (PT) & 0.59195 & 0.58594 & 0.03509 & \textbf{0.67501}\tnote{$\dagger$} \\
       Llama3-1B (IT) & 0.58722 &  0.50072 & \textbf{0.08553}\tnote{$\dagger$} & \textbf{0.67083}\tnote{$\dagger$} \\
       Llama3-3B (IT) & \textbf{0.61592}\tnote{$\dagger$} & 0.52352 & \textbf{0.07061}\tnote{$\dagger$} & \textbf{0.69595}\tnote{$\dagger$} \\
       Llama3-8B (IT) & \textbf{0.61986}\tnote{$\dagger$} & 0.54813 & \textbf{0.05570}\tnote{$\dagger$} & \textbf{0.67503}\tnote{$\dagger$} \\
       Llama3-70B (IT) & 0.59195 & 0.53864 & \textbf{0.06535}\tnote{$\dagger$} & \textbf{0.69817}\tnote{$\dagger$} \\
       \hdashline
       \rowcolor{gray!20} Gemma3-1B (PT) & \textbf{0.63923}\tnote{$\dagger$} & 0.58475 & 0.03421 & \textbf{0.64280}\tnote{$\dagger$} \\
       \rowcolor{gray!20} Gemma3-4B (PT) & \textbf{0.62373}\tnote{$\dagger$} & 0.57981 & 0.03158 & \textbf{0.65521}\tnote{$\dagger$} \\
       \rowcolor{gray!20} Gemma3-12B (PT) & \textbf{0.61104}\tnote{$\dagger$} & 0.58613 & 0.02807 & \textbf{0.66392}\tnote{$\dagger$} \\
       \rowcolor{gray!20} Gemma3-27B (PT) & \textbf{0.62077}\tnote{$\dagger$} & 0.59398 & 0.03114 & \textbf{0.65530}\tnote{$\dagger$} \\
       Gemma3-1B (IT) & \textbf{0.62703}\tnote{$\dagger$} & 0.55655 & \textbf{0.06711}\tnote{$\dagger$} & \textbf{0.70714}\tnote{$\dagger$} \\
       Gemma3-4B (IT) & 0.58937 & 0.51274 & \textbf{0.07061}\tnote{$\dagger$} & \textbf{0.72819}\tnote{$\dagger$} \\
       Gemma3-12B (IT) & 0.59290 & 0.52782 & \textbf{0.08597}\tnote{$\dagger$} & \textbf{0.72619}\tnote{$\dagger$} \\
       Gemma3-27B (IT) & \textbf{0.53554}\tnote{$\dagger$} & 0.46521 & \textbf{0.10219}\tnote{$\dagger$} & \textbf{0.71779}\tnote{$\dagger$} \\
     \toprule
    \end{tabular}
    \caption{Metric scores of common indicators of linguistic convergence in model-generated responses to conversations in the \textit{Movie}s dataset. Scores for \texttt{Utterance Length}, \texttt{PPROPN}, and \texttt{Token Novelty} in \textbf{bold} are significantly different from human metrics ($p<0.05$ on a paired t-test); scores indicated with $\dagger$ are $p < 0.001$ relative to human scores.}
    \label{tab:app-results-movie}
    \end{threeparttable}
\end{table*}

%% file: npr_table.tex
\begin{table*}[]
    \centering
    \begin{threeparttable}
    \begin{tabular}{l r r r r r r}
    \toprule
       \multirow{2}{*}{{Model}} & \multicolumn{4}{c}{{Accommodation Features}} \\
       & Utterance Len. ($\uparrow$) & LIWC ($\uparrow$) & PROPN ($\uparrow$) & Token Novelty ($\downarrow$) \\
       \hline
       Human & 0.45905 & 0.53007 & 0.25837 & 0.58561 \\
       Random & 0.46025 & 0.48035 & 0.00959 & 0.58921 \\
       \arrayrulecolor{black}\hdashline
       \rowcolor{gray!20} Llama3-1B (PT) & \textbf{0.63248}\tnote{$\dagger$} & 0.52466 & \textbf{0.10627}\tnote{$\dagger$} & \textbf{0.45246}\tnote{$\dagger$} \\
       \rowcolor{gray!20} Llama3-3B (PT) & \textbf{0.61824}\tnote{$\dagger$} & 0.52563 & \textbf{0.11510}\tnote{$\dagger$} & \textbf{0.52262}\tnote{$\dagger$} \\
       \rowcolor{gray!20} Llama3-8B (PT) & \textbf{0.62951}\tnote{$\dagger$} & 0.53303 & \textbf{0.11480}\tnote{$\dagger$} & \textbf{0.50684}\tnote{$\dagger$} \\
       \rowcolor{gray!20} Llama3-70B (PT) & \textbf{0.59195}\tnote{$\dagger$} & 0.53057 & \textbf{0.14495}\tnote{$\dagger$} & \textbf{0.49636}\tnote{$\dagger$} \\
       Llama3-1B (IT) & \textbf{0.58722}\tnote{$\dagger$} & 0.52972 & 0.26675 & \textbf{0.60536}\tnote{$\dagger$} \\
       Llama3-3B (IT) & \textbf{0.61592}\tnote{$\dagger$} & 0.52786 & \textbf{0.23097} & \textbf{0.60754}\tnote{$\dagger$} \\
       Llama3-8B (IT) & \textbf{0.61986}\tnote{$\dagger$} & 0.53835 & \textbf{0.21894}\tnote{$\dagger$} & \textbf{0.56910}\tnote{$\dagger$} \\
       Llama3-70B (IT) & \textbf{0.59195}\tnote{$\dagger$} & 0.53203 & 0.26249 & \textbf{0.59674}\tnote{$\dagger$} \\
       \hdashline
       \rowcolor{gray!20} Gemma3-1B (PT) & \textbf{0.63923}\tnote{$\dagger$} & 0.51606 & \textbf{0.12287}\tnote{$\dagger$} & \textbf{0.44413}\tnote{$\dagger$} \\
       \rowcolor{gray!20} Gemma3-4B (PT) & \textbf{0.62373}\tnote{$\dagger$} & 0.52087 & \textbf{0.11754}\tnote{$\dagger$} & \textbf{0.50963}\tnote{$\dagger$} \\
       \rowcolor{gray!20} Gemma3-12B (PT) & \textbf{0.61104}\tnote{$\dagger$} & 0.52570 & \textbf{0.12500}\tnote{$\dagger$} & \textbf{0.51644}\tnote{$\dagger$} \\
       \rowcolor{gray!20} Gemma3-27B (PT) & \textbf{0.62077}\tnote{$\dagger$} & 0.52608 & \textbf{0.12180}\tnote{$\dagger$} & \textbf{0.51689}\tnote{$\dagger$} \\
       Gemma3-1B (IT) & \textbf{0.62703}\tnote{$\dagger$} & 0.52380 & \textbf{0.16261}\tnote{$\dagger$} & \textbf{0.64512}\tnote{$\dagger$} \\
       Gemma3-4B (IT) & \textbf{0.58937}\tnote{$\dagger$} & 0.52234 & \textbf{0.15164}\tnote{$\dagger$} & \textbf{0.67585}\tnote{$\dagger$} \\
       Gemma3-12B (IT) & \textbf{0.59290}\tnote{$\dagger$} & 0.52603 & \textbf{0.17266}\tnote{$\dagger$} & \textbf{0.65817}\tnote{$\dagger$} \\
       Gemma3-27B (IT) & \textbf{0.53554}\tnote{$\dagger$} & 0.51925 & \textbf{0.16078}\tnote{$\dagger$} & \textbf{0.68015}\tnote{$\dagger$} \\
     \toprule
    \end{tabular}
    \caption{Metric scores of common indicators of linguistic convergence in model-generated responses to conversations in the \textit{NPR} Interview dataset. Scores for \texttt{Utterance Length}, \texttt{PPROPN}, and \texttt{Token Novelty} in \textbf{bold} are significantly different from human metrics ($p<0.05$ on a paired t-test); scores indicated with $\dagger$ are $p < 0.001$ relative to human scores.}
    \label{tab:app-results-npr}
    \end{threeparttable}
\end{table*}

%% file: liwc_dailydialog_table.tex
\begin{sidewaystable*}[]
    \centering
    \begin{threeparttable}
    \begin{tabular}{l r r r r r r r r r}
    \toprule
       \multirow{2}{*}{{Model}} & \multicolumn{9}{c}{{LIWC Classes} $\uparrow$} \\
       & \texttt{Per. PRON} & \texttt{Imp. PRON} & \texttt{Article} & \texttt{CONJ} & \texttt{PREP} & \texttt{AUX Verb} & \texttt{Adverb} & \texttt{Negation} & \texttt{Quanitifer} \\
       \hline
       Human & 0.55537 & 0.47056 & 0.50837 & 0.499783 & 0.47303 & 0.45335 & 0.46373 & 0.82350 & 0.59065 \\
       Random & 0.49129 & 0.42833 & 0.43827 & 0.49319 & 0.41357 & 0.42089 & 0.45927 & 0.81013 & 0.58833 \\
       \arrayrulecolor{black}\hdashline
       \rowcolor{gray!20} Llama3-1B (PT) & 0.59106\tnote{\textbf{+}} & 0.53085\tnote{\textbf{+}} & 0.54780\tnote{\textbf{+}} & 0.57325\tnote{\textbf{+}}& 0.48835 & 0.46233 & 0.51702\tnote{\textbf{+}}& 0.79396\tnote{\textbf{-}} & 0.70274\tnote{\textbf{+}} \\
       \rowcolor{gray!20} Llama3-3B (PT) & 0.57755\tnote{\textbf{+}} & 0.52371\tnote{\textbf{+}} & 0.53142 & 0.55528\tnote{\textbf{+}} & 0.46733 & 0.47396 & 0.52381\tnote{\textbf{+}} & 0.81289 & 0.69351\tnote{\textbf{+}} \\
       \rowcolor{gray!20} Llama3-8B (PT) & 0.58959\tnote{\textbf{+}} & 0.51368\tnote{\textbf{+}} & 0.52987 & 0.54597\tnote{\textbf{+}} & 0.47035 & 0.48503\tnote{\textbf{+}} & 0.51281\tnote{\textbf{+}} & 0.84558\tnote{\textbf{+}} & 0.69377\tnote{\textbf{+}} \\
       \rowcolor{gray!20} Llama3-70B (PT) & 0.58169\tnote{\textbf{+}} & 0.50944\tnote{\textbf{+}} & 0.52370 & 0.56084\tnote{\textbf{+}} & 0.47408 & 0.45810 & 0.51558\tnote{\textbf{+}} & 0.82351 & 0.69056\tnote{\textbf{+}} \\
       Llama3-1B (IT) & 0.60780\tnote{\textbf{+}} & 0.47280 & 0.44638\tnote{\textbf{-}} & 0.48232 & 0.48471 & 0.47022 & 0.48627 & 0.79248\tnote{\textbf{-}} & 0.61341 \\
       Llama3-3B (IT) & 0.58567\tnote{\textbf{+}} & 0.47435 & 0.49488 & 0.47292\tnote{\textbf{-}} & 0.47582 & 0.48030\tnote{\textbf{+}} & 0.47990 & 0.82840 & 0.62471\tnote{\textbf{+}} \\
       Llama3-8B (IT) & 0.59796\tnote{\textbf{+}} & 0.46956 & 0.48389 & 0.50265 & 0.48433 & 0.48032\tnote{\textbf{+}} & 0.48493 & 0.83320 & 0.63653\tnote{\textbf{+}} \\
       Llama3-70B (IT) & 0.57476 & 0.45832 & 0.45687\tnote{\textbf{-}} & 0.47638 & 0.46646 & 0.45063 & 0.44361 & 0.81702 & 0.60406 \\
       \hdashline
       \rowcolor{gray!20} Gemma3-1B (PT) & 0.60207\tnote{\textbf{+}} & 0.53798\tnote{\textbf{+}} & 0.53470\tnote{\textbf{+}} & 0.56730\tnote{\textbf{+}} & 0.50729\tnote{\textbf{+}} & 0.47569 & 0.50796\tnote{\textbf{+}} & 0.78080\tnote{\textbf{-}} & 0.69818\tnote{\textbf{+}} \\
       \rowcolor{gray!20} Gemma3-4B (PT) & 0.57737\tnote{\textbf{+}} & 0.52909\tnote{\textbf{+}} & 0.53040 & 0.57401\tnote{\textbf{+}} & 0.46686 & 0.45656 & 0.52408\tnote{\textbf{+}} & 0.81130 & 0.70267\tnote{\textbf{+}} \\
       \rowcolor{gray!20} Gemma3-12B (PT) & 0.57765\tnote{\textbf{+}} & 0.51877\tnote{\textbf{+}} & 0.53791\tnote{\textbf{+}} & 0.55138\tnote{\textbf{+}} & 0.48060 & 0.45863 & 0.51551\tnote{\textbf{+}} & 0.83322 & 0.68864\tnote{\textbf{+}} \\
       \rowcolor{gray!20} Gemma3-27B (PT) & 0.56882 & 0.51818\tnote{\textbf{+}} & 0.52879 & 0.54490\tnote{\textbf{+}} & 0.46479 & 0.44267 & 0.51152\tnote{\textbf{+}} & 0.82200 & 0.68911\tnote{\textbf{+}} \\
       Gemma3-1B (IT) & 0.58208\tnote{\textbf{+}} & 0.49887\tnote{\textbf{+}} & 0.50235 & 0.54368\tnote{\textbf{+}} & 0.49011 & 0.46066 & 0.51424\tnote{\textbf{+}} & 0.85839\tnote{\textbf{+}} & 0.67683\tnote{\textbf{+}} \\
       Gemma3-4B (IT) & 0.55696 & 0.42369\tnote{\textbf{-}} & 0.43792\tnote{\textbf{-}} & 0.43233\tnote{\textbf{-}} & 0.45948 & 0.45446 & 0.45307 & 0.85591\tnote{\textbf{+}} & 0.59141 \\
       Gemma3-12B (IT) & 0.55628 & 0.43151\tnote{\textbf{-}} & 0.45384\tnote{\textbf{-}} & 0.45801 & 0.46614\tnote{\textbf{-}} & 0.43746 & 0.44574 & 0.83848 & 0.57419 \\
       Gemma3-27B (IT) & 0.53463 & 0.41544 & 0.42213\tnote{\textbf{-}} & 0.40541\tnote{\textbf{-}} & 0.44547\tnote{\textbf{-}} & 0.44339 & 0.41977\tnote{\textbf{-}} & 0.81694 & 0.53432\tnote{\textbf{-}} \\
     \toprule
    \end{tabular}
    \caption{Per-class scores of LIWC categories on the \textit{DailyDialog} dataset. Per-class scores significantly (p<0.05) over- or under-accommodating relative to the human baseline are annotated with $^{+}/^{-}$, respectively.}
    \label{tab:liwc-results-dailydialog}
    \end{threeparttable}
\end{sidewaystable*}

%% file: liwc_movie_table.tex
\begin{sidewaystable*}[]
    \centering
    \begin{threeparttable}
    \begin{tabular}{l r r r r r r r r r}
    \toprule
       \multirow{2}{*}{{Model}} & \multicolumn{9}{c}{{LIWC Classes} $\uparrow$} \\
       & \texttt{Per. PRON} & \texttt{Imp. PRON} & \texttt{Article} & \texttt{CONJ} & \texttt{PREP} & \texttt{AUX Verb} & \texttt{Adverb} & \texttt{Negation} & \texttt{Quanitifer} \\
       \hline
       Human & 0.48662 & 0.45882 & 0.51208 & 0.53520 & 0.42770 & 0.44849 & 0.49068 & 0.73628 & 0.69347 \\
       Random & 0.47229 & 0.43810 & 0.49400 & 0.54280 & 0.43686 & 0.41626 & 0.48199 & 0.73486 & 0.68726 \\
       \arrayrulecolor{black}\hdashline
       \rowcolor{gray!20} Llama3-1B (PT) & 0.56046\tnote{\textbf{+}} & 0.51947\tnote{\textbf{+}} & 0.56849\tnote{\textbf{+}} & 0.61117\tnote{\textbf{+}} & 0.49737\tnote{\textbf{+}} & 0.47484\tnote{\textbf{+}} & 0.56384\tnote{\textbf{+}} & 0.66310\tnote{\textbf{-}} & 0.78845\tnote{\textbf{+}} \\
       \rowcolor{gray!20} Llama3-3B (PT) & 0.55145\tnote{\textbf{+}} & 0.50901\tnote{\textbf{+}} & 0.56037\tnote{\textbf{+}} & 0.63205\tnote{\textbf{+}} & 0.50568\tnote{\textbf{+}} & 0.46281 & 0.58514\tnote{\textbf{+}} & 0.69707\tnote{\textbf{-}} & 0.78515\tnote{\textbf{+}} \\
       \rowcolor{gray!20} Llama3-8B (PT) & 0.55398\tnote{\textbf{+}} & 0.51839\tnote{\textbf{+}} & 0.57427\tnote{\textbf{+}} & 0.63041\tnote{\textbf{+}} & 0.51024\tnote{\textbf{+}} & 0.47168 & 0.58988\tnote{\textbf{+}} & 0.66916\tnote{\textbf{-}} & 0.78227\tnote{\textbf{+}} \\
       \rowcolor{gray!20} Llama3-70B (PT) & 0.54275\tnote{\textbf{+}} & 0.50662\tnote{\textbf{+}} & 0.57818\tnote{\textbf{+}} & 0.62881\tnote{\textbf{+}} & 0.51181\tnote{\textbf{+}} & 0.45624 & 0.58017\tnote{\textbf{+}} & 0.68024\tnote{\textbf{-}} & 0.78867\tnote{\textbf{+}} \\
       Llama3-1B (IT) & 0.57487\tnote{\textbf{+}} & 0.44679\tnote{\textbf{-}} & 0.43215 & 0.51564 & 0.43272 & 0.43078 & 0.46555 & 0.54436\tnote{\textbf{-}} & 0.66362\tnote{\textbf{-}} \\
       Llama3-3B (IT) & 0.55139\tnote{\textbf{+}} & 0.44934 & 0.48609\tnote{\textbf{-}} & 0.51258 & 0.44856 & 0.44119 & 0.48063 & 0.64849\tnote{\textbf{-}} & 0.69343 \\
       Llama3-8B (IT) & 0.54993\tnote{\textbf{+}} & 0.46742 & 0.49897 & 0.56659\tnote{\textbf{+}} & 0.47575\tnote{\textbf{+}} & 0.46808 & 0.52338\tnote{\textbf{+}} & 0.67381\tnote{\textbf{-}} & 0.70928 \\
       Llama3-70B (IT) & 0.54230\tnote{\textbf{+}} & 0.45001 & 0.50772 & 0.54154 & 0.45813\tnote{\textbf{+}} & 0.47079 & 0.50060 & 0.69068\tnote{\textbf{-}} & 0.68594 \\
       \hdashline
       \rowcolor{gray!20} Gemma3-1B (PT) & 0.57242\tnote{\textbf{+}} & 0.54646\tnote{\textbf{+}} & 0.54701\tnote{\textbf{+}} & 0.62979\tnote{\textbf{+}} & 0.51441\tnote{\textbf{+}} & 0.46974 & 0.58130\tnote{\textbf{+}} & 0.60856\tnote{\textbf{-}} & 0.79308\tnote{\textbf{+}} \\
       \rowcolor{gray!20} Gemma3-4B (PT) & 0.55490\tnote{\textbf{+}} & 0.51745\tnote{\textbf{+}} & 0.56900\tnote{\textbf{+}} & 0.62117\tnote{\textbf{+}} & 0.50467\tnote{\textbf{+}} & 0.46693 & 0.58258\tnote{\textbf{+}} & 0.61537\tnote{\textbf{-}} & 0.78623\tnote{\textbf{+}} \\
       \rowcolor{gray!20} Gemma3-12B (PT) & 0.55298\tnote{\textbf{+}} & 0.51403\tnote{\textbf{+}} & 0.57491\tnote{\textbf{+}} & 0.62396\tnote{\textbf{+}} & 0.50916\tnote{\textbf{+}} & 0.46872 & 0.58389\tnote{\textbf{+}} & 0.66710\tnote{\textbf{-}} & 0.78044\tnote{\textbf{+}} \\
       \rowcolor{gray!20} Gemma3-27B (PT) & 0.55451\tnote{\textbf{+}} & 0.53745\tnote{\textbf{+}} & 0.59499\tnote{\textbf{+}} & 0.63705\tnote{\textbf{+}} & 0.50941\tnote{\textbf{+}} & 0.49088\tnote{\textbf{+}} & 0.58387\tnote{\textbf{+}} & 0.64797\tnote{\textbf{-}} & 0.78968\tnote{\textbf{+}} \\
       Gemma3-1B (IT) & 0.51589\tnote{\textbf{+}} & 0.47635 & 0.48902 & 0.60815\tnote{\textbf{+}} & 0.45868\tnote{\textbf{+}} & 0.46234 & 0.53365\tnote{\textbf{+}} & 0.71364\tnote{\textbf{-}} & 0.75127\tnote{\textbf{+}} \\
       Gemma3-4B (IT) & 0.47517 & 0.42279\tnote{\textbf{-}} & 0.44448\tnote{\textbf{-}} & 0.52310 & 0.42822 & 0.42271\tnote{\textbf{-}} & 0.45371\tnote{\textbf{-}} & 0.74263 & 0.70185 \\
       Gemma3-12B (IT) & 0.51448\tnote{\textbf{+}} & 0.43624 & 0.49435 & 0.52473 & 0.45220 & 0.44352 & 0.48475 & 0.71455\tnote{\textbf{-}} & 0.68557 \\
       Gemma3-27B (IT) & 0.48289 & 0.36519\tnote{\textbf{-}} & 0.41157\tnote{\textbf{-}} & 0.42902\tnote{\textbf{-}} & 0.39545\tnote{\textbf{-}} & 0.38945\tnote{\textbf{-}} & 0.38815\tnote{\textbf{-}} & 0.68913\tnote{\textbf{-}} & 0.63603\tnote{\textbf{-}} \\
     \toprule
    \end{tabular}
    \caption{Per-class scores of LIWC categories on the \textit{Movie} corpus. Per-class scores significantly (p<0.05) over- or under-accommodating relative to the human baseline are annotated with $^{+}/^{-}$, respectively.}
    \label{tab:liwc-results-movie}
    \end{threeparttable}
\end{sidewaystable*}

%% file: liwc_npr_table.tex
\begin{sidewaystable*}[]
    \centering
    \begin{threeparttable}
    \begin{tabular}{l r r r r r r r r r}
    \toprule
       \multirow{2}{*}{{Model}} & \multicolumn{9}{c}{{LIWC Classes} $\uparrow$} \\
       & \texttt{Per. PRON} & \texttt{Imp. PRON} & \texttt{Article} & \texttt{CONJ} & \texttt{PREP} & \texttt{AUX Verb} & \texttt{Adverb} & \texttt{Negation} & \texttt{Quanitifer} \\
       \hline
       Human & 0.51224 & 0.56815 & 0.52201 & 0.50898 & 0.57620 & 0.58099 & 0.46672 & 0.62373 & 0.41161 \\
       Random & 0.47656 & 0.46571 & 0.48411 & 0.46186 & 0.55681 & 0.49711 & 0.44354 & 0.53601 & 0.40152 \\
       \arrayrulecolor{black}\hdashline
       \rowcolor{gray!20} Llama3-1B (PT) & 0.51461 & 0.54908\tnote{\textbf{-}} & 0.50790\tnote{\textbf{-}} & 0.41766\tnote{\textbf{-}} & 0.44772\tnote{\textbf{-}} & 0.48655\tnote{\textbf{-}} & 0.46538 & 0.80887\tnote{\textbf{+}} & 0.52413\tnote{\textbf{+}} \\
       \rowcolor{gray!20} Llama3-3B (PT) & 0.50221 & 0.54298\tnote{\textbf{-}} & 0.50162\tnote{\textbf{-}} & 0.42035\tnote{\textbf{-}} & 0.45789\tnote{\textbf{-}} & 0.50929\tnote{\textbf{-}} & 0.47096 & 0.81561\tnote{\textbf{+}} & 0.50974\tnote{\textbf{+}} \\
       \rowcolor{gray!20} Llama3-8B (PT) & 0.50871 & 0.55488\tnote{\textbf{-}} & 0.50834\tnote{\textbf{-}} & 0.41920\tnote{\textbf{-}} & 0.47860\tnote{\textbf{-}} & 0.52426\tnote{\textbf{-}} & 0.47534 & 0.80852\tnote{\textbf{+}} & 0.51937\tnote{\textbf{+}} \\
       \rowcolor{gray!20} Llama3-70B (PT) & 0.51743 & 0.53373\tnote{\textbf{-}} & 0.50334\tnote{\textbf{-}} & 0.42873\tnote{\textbf{-}} & 0.49814\tnote{\textbf{-}} & 0.50906\tnote{\textbf{-}} & 0.46667 & 0.78768\tnote{\textbf{+}} & 0.53037\tnote{\textbf{+}} \\
       Llama3-1B (IT) & 0.51252 & 0.52152\tnote{\textbf{-}} & 0.51547 & 0.46362\tnote{\textbf{-}} & 0.56144\tnote{\textbf{-}} & 0.52774\tnote{\textbf{-}} & 0.42933\tnote{\textbf{-}} & 0.77361\tnote{\textbf{+}} & 0.46223\tnote{\textbf{+}} \\
       Llama3-3B (IT) & 0.50690 & 0.52605\tnote{\textbf{-}} & 0.52064 & 0.44533\tnote{\textbf{-}} & 0.54712\tnote{\textbf{-}} & 0.52491\tnote{\textbf{-}} & 0.43244\tnote{\textbf{-}} & 0.78154\tnote{\textbf{+}} & 0.46584\tnote{\textbf{+}} \\
       Llama3-8B (IT) & 0.52654\tnote{\textbf{+}} & 0.54591\tnote{\textbf{-}} & 0.51884 & 0.44775\tnote{\textbf{-}} & 0.53512\tnote{\textbf{-}} & 0.55602\tnote{\textbf{-}} & 0.43379\tnote{\textbf{-}} & 0.79366\tnote{\textbf{+}} & 0.48750\tnote{\textbf{+}} \\
       Llama3-70B (IT) & 0.52543\tnote{\textbf{+}} & 0.52190\tnote{\textbf{-}} & 0.51127 & 0.44526\tnote{\textbf{-}} & 0.53883\tnote{\textbf{-}} & 0.53572\tnote{\textbf{-}} & 0.43549\tnote{\textbf{-}} & 0.79175\tnote{\textbf{+}} & 0.48261\tnote{\textbf{+}} \\
       \hdashline
       \rowcolor{gray!20} Gemma3-1B (PT) & 0.50781 & 0.53966\tnote{\textbf{-}} & 0.48955\tnote{\textbf{-}} & 0.40049\tnote{\textbf{-}} & 0.46065\tnote{\textbf{-}} & 0.48000\tnote{\textbf{-}} & 0.46200 & 0.78614\tnote{\textbf{+}} & 0.51833\tnote{\textbf{+}} \\
       \rowcolor{gray!20} Gemma3-4B (PT) & 0.49885\tnote{\textbf{-}} & 0.54071\tnote{\textbf{-}} & 0.49152\tnote{\textbf{-}} & 0.41224\tnote{\textbf{-}} & 0.46063\tnote{\textbf{-}} & 0.50473\tnote{\textbf{-}} & 0.46580 & 0.79908\tnote{\textbf{+}} & 0.51428\tnote{\textbf{+}} \\
       \rowcolor{gray!20} Gemma3-12B (PT) & 0.49863\tnote{\textbf{-}} & 0.54130\tnote{\textbf{-}} & 0.49419\tnote{\textbf{-}} & 0.41503\tnote{\textbf{-}} & 0.47925\tnote{\textbf{-}} & 0.51447\tnote{\textbf{-}} & 0.46118 & 0.80888\tnote{\textbf{+}} & 0.51833\tnote{\textbf{+}} \\
       \rowcolor{gray!20} Gemma3-27B (PT) & 0.50531 & 0.54580\tnote{\textbf{-}} & 0.49988\tnote{\textbf{-}} & 0.41498\tnote{\textbf{-}} & 0.48409\tnote{\textbf{-}} & 0.51519\tnote{\textbf{-}} & 0.46871 & 0.78867\tnote{\textbf{+}} & 0.51207\tnote{\textbf{+}} \\
       Gemma3-1B (IT) & 0.46099\tnote{\textbf{-}} & 0.54123\tnote{\textbf{-}} & 0.51009\tnote{\textbf{-}} & 0.45534\tnote{\textbf{-}} & 0.49789\tnote{\textbf{-}} & 0.47316\tnote{\textbf{-}} & 0.44876\tnote{\textbf{-}} & 0.82397\tnote{\textbf{+}} & 0.50275\tnote{\textbf{+}} \\
       Gemma3-4B (IT) & 0.47673\tnote{\textbf{-}} & 0.53623\tnote{\textbf{-}} & 0.51586 & 0.44279\tnote{\textbf{-}} & 0.50731\tnote{\textbf{-}} & 0.48312\tnote{\textbf{-}} & 0.45753 & 0.80174\tnote{\textbf{+}} & 0.47973\tnote{\textbf{+}} \\
       Gemma3-12B (IT) & 0.47781\tnote{\textbf{-}} & 0.52282\tnote{\textbf{-}} & 0.51185 & 0.45791\tnote{\textbf{-}} & 0.54708\tnote{\textbf{-}} & 0.50205\tnote{\textbf{-}} & 0.44337\tnote{\textbf{-}} & 0.81047\tnote{\textbf{+}} & 0.46087\tnote{\textbf{+}} \\
       Gemma3-27B (IT) & 0.48005\tnote{\textbf{-}} & 0.50935\tnote{\textbf{-}} & 0.49790\tnote{\textbf{-}} & 0.44714\tnote{\textbf{-}} & 0.54294\tnote{\textbf{-}} & 0.47903\tnote{\textbf{-}} & 0.45856 & 0.80388\tnote{\textbf{+}} & 0.45444\tnote{\textbf{+}} \\
     \toprule
    \end{tabular}
    \caption{Per-class scores of LIWC categories on the \textit{NPR} Interview corpus. Per-class scores significantly (p<0.05) over- or under-accommodating relative to the human baseline are annotated with $^{+}/^{-}$, respectively.}
    \label{tab:liwc-results-npr}
    \end{threeparttable}
\end{sidewaystable*}